# Dual Classification Head Self-training Network for Cross-scene Hyperspectral Image Classification

Rong Liu, *Member, IEEE*, Junye Liang, Jiaqi Yang, Jiang He, and Peng Zhu.

*Abstract*—Due to the difficulty of obtaining labeled data for hyperspectral images (HSIs), cross-scene classification has emerged as a widely adopted approach in the remote sensing community. It involves training a model using labeled data from a source domain (SD) and unlabeled data from a target domain (TD), followed by inferencing on the TD. However, variations in the reflectance spectrum of the same object between the SD and the TD, as well as differences in the feature distribution of the same land cover class, pose significant challenges to the performance of cross-scene classification. To address this issue, we propose a dual classification head self-training network (DHSNet). This method aligns class-wise features across domains, ensuring that the trained classifier can accurately classify TD data of different classes. We introduce a dual classification head self-training strategy for the first time in the cross-scene HSI classification field. The proposed approach mitigates domain gap while preventing the accumulation of incorrect pseudo-labels in the model. Additionally, we incorporate a novel central feature attention mechanism to enhance the model's capacity to learn scene-invariant features across domains. Experimental results on three cross-scene HSI datasets demonstrate that the proposed DHSNet significantly outperforms other state-of-the-art approaches. The code for DHSNet will be available at https://github.com/liurongwhm.

## I. INTRODUCTION

With the advancements in sensor technology, aerospace engineering and computer science, hyperspectral image classification has become a prominent research topic in the field of remote sensing [1], [2]. It aims at assigning each pixel to a predefined category. Hyperspectral images (HSIs), known for their high spectral resolution, provide strong technique support for precise land cover classification and are widely applied in various fields, such as agricultural monitoring [3], geological exploration [4], environmental monitoring [5], and smart city development [6].

HSI classification faces significant challenges due to limited labels and high dimensionality [7]. Over the past few decades, HSI classification methods for the same scene have been extensively studied [8-12], generally based on the assumption that the training and test data are independent and identically distributed. While this assumption roughly holds within the same scene, enabling these methods to achieve remarkable success, and it breaks down in cross-scene scenarios [13]. Due to the labeling burden and expensive time cost [14], the HSI used for model training (source domain, SD) and the HSI to be classified (target domain, TD) are often from different scenes. As a result, applying models trained in one region to another region becomes crucial for practical applications, making cross-domain classification a key focus when classifying HSIs. Due to the variations in sensor conditions, geographical locations, and environmental factors, SD and TD have huge differences in spectral reflectance and feature distribution, even for the same land cover class [15]. The above phenomenon results in a domain shift, where the distribution mismatch between the two domains leads to a substantial drop of classification performance.

To address the challenge of model transferability across different scenes, two major types of methods from transfer learning are developed: domain adaptation (DA) and domain generalization (DG). Domain adaptation focuses on adapting models trained on SD to perform well on a specific TD with access to TD data and even partial TD labels. The ultimate object of DA is to overcome domain shift and transfer invariant knowledge from the SD to the TD. Domain generalization aims to create models capable of generalizing to TD without access to TD data during training. They focus on architectural designs aimed at enhancing cross-domain generalization. Although a limited number of studies have delved into the domain generalization of cross-scene HSI classification and have achieved commendable advancements [16-18], DG-based approaches necessitate the generation and discrimination of expanded domain samples. These approaches are characterized by a substantial requirement for SD samples and are often encumbered by significant computational overhead. Consequently, research and application of DA-based methods have been more extensively pursued and are deemed more efficient.

Traditional DA methods can be categorized into two main types [19]: instance-based [20-22] and feature-based methods [23-25]. Instance-based methods focus on minimizing domain discrepancies by reweighting SD samples and training models on these adjusted datasets. Feature-based methods, on the other hand, aim to construct a shared feature space where the distributions of SD and TD can be aligned. However,

This work was supported in part by the National Natural Science Foundation of China under Grant 62201622. (*Corresponding author: Jiaqi Yang*).

Rong Liu and Junye Liang are with the School of Geography and Planning, Sun Yat-Sen University, Guangzhou 510275, China (e-mail: liurong25@mail.sysu.edu.cn; liangjy225@mail2.sysu.edu.cn).

Jiaqi Yang is with Department of Forest and Wildlife Ecology, University of Wisconsin-Madison, 1630 Linden Dr., Madison, WI 53706, USA (e-mail: jiaqi.yang@wisc.edu).

Jiang He is with Chair of Data Science in Earth Observation, Technical University of Munich, Munich, 80333, Germany (e-mail: jiang.he@tum.de).

Peng Zhu is with Department of Geography and Institute for Climate and Carbon Neutrality, The University of Hong Kong, Hong Kong SAR, China (email: zhupeng@hku.hk).

traditional DA methods face difficulties in feature construction and often fail to obtain transferable features.

The ability of deep learning to extract high-level and abstract features has led to the promotion of the flourishing of DA, which can be divided into discrepancy-based methods and adversarial-based methods [26]. Discrepancy-based methods explicitly measure the difference in feature distributions between SD and TD. A common metric is maximum mean discrepancy (MMD) [27], which maps data from both domains to a reproducing kernel Hilbert space and computes the distance between their averages as a loss function. Based on this, Long et al. [28] introduced the deep adaptation network (DAN), which employs multi-kernel MMD (MK-MMD), while their later work on joint adaptation networks (JAN) [29] introduced joint MMD (JMMD) to align joint distributions across multiple layers. Besides MMD, there are other discrepancy metrics. CORAL is defined as the distance between the covariances of the source and target features [30] It was first introduced into deep learning by Sun et al. [31]. Optimal transport [32] (OT) is a theory that allows to compare probability distributions in a geometrically sound manner. Damodaran et al. proposed the DeepJDOT by minimizing the discrepancy of joint domain distributions by OT [33]. Liu et al. propose a method integrates OT theory with clustering operation, termed Clustering-based Optimal Transport (COT) [34].

Adversarial-based methods implicitly align cross-domain features through adversarial learning [35]. For example, domain-adversarial neural networks (DANNs) [36], inspired by generative adversarial networks (GANs) [37], employ a domain discriminator to distinguish whether a data point originates from SD or TD. The model is trained in an adversarial manner to confuse the discriminator, aiming to minimize the feature distribution differences between the two domains. As a result, the model, despite being trained on SD, can generalize to TD because the domain classifier is unable to distinguish between them. Beyond using a single domain discriminator, other studies explore alternative approaches. Methods such as [38-40] employ two classifiers as discriminators to maximize the discrepancy between their predictions, while the feature extractor is simultaneously trained to minimize this discrepancy. This adversarial process helps bridge the gap between SD and TD by making the learned features transferable across domains, effectively reducing domain shift.

Deep DA methods have shown effectiveness in cross-scene HSI classification [41]. However, the extent and characteristics of data distribution shifts differ across land cover classes, leading to distinct patterns in the joint distributions of domain and class. To address this issue, several methods have been proposed. Liu et al. [42] proposed a cross-scene HSI classification method that aligns the conditional distribution of each class by combining class-wise DANN and MMD, using pseudo-labels of target data during model optimization. Zhang et al. [43] proposed a topological structure and semantic information transfer network (TSTnet), which aligns both the statistical and geometric distributions between SD and TD. Fang et al. [44] developed a confident learning-based bi-classifier adversarial neural network (CLDA) with a dynamic distribution adaptation strategy, dynamically adjusting the importance of marginal and conditional distributions. Zhao et al. [45] presented a framework toward multilevel features and decision boundaries (ToMF-B), which simultaneously aligns task-related features and learns task-specific decision boundaries. Li et al. [46] proposed a dual-channel residual network with a dynamic distribution adaptation strategy (DDAN-JCA). These methods have incorporated the characteristics of hyperspectral imagery into the development of DA models, achieving notable results.

In most existing models, pseudo-supervisory information from the TD directly influences the optimization of overall model parameters. It often leads to incorrect judgments that can negatively impact subsequent training and cause unpredictable performance degradation. Additionally, most existing methods for cross-scene HSI classification focus on framework design, ignoring feature extractors that can represent cross-scene invariant features. To address these limitations, we propose a dual classification head self-training network. This method introduces a novel central feature attention mechanism during feature extraction and employs class-wise feature alignment to align features between SD and TD. Furthermore, it uses dual classification heads for self-training, reducing model bias towards the SD while preventing the accumulation of incorrect pseudo-labels in the model.

The main contributions of our proposed method are summarized as follows:

1) A novel cross-scene HSI classification framework is proposed. This framework consists of an effective feature extractor and two classification heads. Class-wise feature alignment is employed after feature extraction. The classification heads are used for guiding the predictions of the other pseudo classification head.

2) A novel Central Feature Attention Aware Convolution (CFAAC) block is proposed, which directs strengthen the attention to central features critical for capturing cross-scene invariant information. By focusing on these key features, the model's capacity to represent generalizable features is enhanced, making it suitable for cross-scene classification tasks.

3) To the best of our knowledge, this is the first time to employ a dual classification head self-training strategy in the HSI cross-scene classification. It mitigates the impact of incorrect pseudo-supervisory information from the TD, ensuring the effectiveness of the self-training and the robustness of the whole model.

4) In comparison to state-of-the-art approaches, our method has achieved significant improvements in classification performance on three benchmark datasets. Notably, on the HyRANK dataset including complex scenes, our method achieves over 10.00% improvement in overall accuracy compared to the average of other methods under the same experimental conditions.

The remainder of this paper is organized as follows: Section II reviews related works on DA and self-training techniques. Section III elucidates the overall framework and detailed implementation of the proposed method. Extensive experiments and their analyses are presented in Section IV. Finally, conclusions are drawn in Section V.

## II. Related Works

### A. Local Maximum Mean Discrepancy

Measuring the distance between two distributions is essential for aligning the source and target domain distributions. Bernhard Schölkopf et al. [27] proposed a statistical method MMD to test whether two distributions, $p$ and $q$, are different based on samples drawn from each. MMD quickly became one of the most widely used techniques for quantifying distribution differences between domains. The MMD distance is defined as:

$$\text{MMD}(\mathcal{F}, p, q) := \sup_{f \in \mathcal{F}} \left( \mathbf{E}_{\mathbf{x}^s \sim p}[f(\mathbf{x}^s)] - \mathbf{E}_{\mathbf{x}^t \sim q}[f(\mathbf{x}^t)] \right) \tag{1}$$

where $\mathcal{F}$ represents a class of functions $f: \mathbb{R}^d \to \mathbb{R}$, with $d$ being the dimension of the data, and $\mathbf{x}^s$ and $\mathbf{x}^t$ denote data from the SD and TD, respectively. The operator $sup(\cdot)$ denotes the supremum. Specifically, MMD is defined as the supremum of the difference in expectations between two distributions after they are mapped to real numbers through the functions in $\mathcal{F}$.

As we can only observe a limited number of samples from the distributions, a biased empirical estimate of the MMD is often more practical. This empirical estimate can be formulated as:

$$\text{MMD}_b(\mathcal{F}, \mathcal{D}_s, \mathcal{D}_t) := \sup_{f \in \mathcal{F}} \left( \frac{1}{n_s} \sum_{i=1}^{n_s} f(\mathbf{x}_i^s) - \frac{1}{n_t} \sum_{j=1}^{n_t} f(\mathbf{x}_j^t) \right) \tag{2}$$

where $\mathcal{D}_s$ and $\mathcal{D}_t$ represents the SD and TD, respectively. $\{\mathbf{x}_i^s\}_{i=1}^{n_s}$ and $\{\mathbf{x}_j^t\}_{j=1}^{n_t}$ are samples from $\mathcal{D}_s$ and $\mathcal{D}_t$, respectively.

If $\mathcal{F}$ is the unit ball in a reproducing kernel Hilbert space (RKHS), the empirical MMD can be efficiently computed as:

$$\text{MMD}(\mathcal{D}_s, \mathcal{D}_t) = \left\| \frac{1}{n_s} \sum_{i=1}^{n_s} \phi(\mathbf{x}_i^s) - \frac{1}{n_t} \sum_{j=1}^{n_t} \phi(\mathbf{x}_j^t) \right\|_{\mathcal{H}}^2 \tag{3}$$

where $\mathcal{H}$ is the RKHS endowed with a characteristic kernel, and, $\phi(\cdot)$ denotes the feature mapping that transforms the original samples into the RKHS. By square expansion of the formula, the inner product in the RKHS space can be converted into a kernel function, allowing the MMD to be calculated directly through the kernel function.

Based on this, Ghifary et al. [47] introduced the MMD metric into feedforward neural networks, pioneering its application in deep transfer learning. This approach has since gained widespread application. Zhu et al. [48] proposed the deep subdomain adaption network (DSAN), which introduced the local maximum mean discrepancy (LMMD) to align features between SD and TD, taking subdomains into consideration. LMMD can be formulated as:

$$\text{LMMD}(\mathcal{D}_s, \mathcal{D}_t) = \frac{1}{C} \sum_{c=1}^{C} \left\| \sum_{i=1}^{n_s} w_i^{sc} \phi(\mathbf{x}_i^s) - \sum_{j=1}^{n_t} w_j^{tc} \phi(\mathbf{x}_j^t) \right\|_{\mathcal{H}}^2 \tag{4}$$

where $w_i^{sc}$ and $w_j^{tc}$ denote the weights of $\mathbf{x}_i^s$ and $\mathbf{x}_j^t$ belonging to class $c$, respectively, and $C$ is the total number of categories. Note that $\sum_{i=1}^{m} w_i^{sc} = 1$ and $\sum_{j=1}^{n} w_j^{tc} = 1$. The weights $w_i^c$ for the sample $\mathbf{x}_i$ from SD and TD are computed as:

$$w_i^{sc} = \frac{y_{ic}^s}{\sum_{j=1}^{n_s} y_{jc}^s} \tag{5}$$

$$w_i^{tc} = \frac{y_{ic}^t}{\sum_{j=1}^{n_t} y_{jc}^t} \tag{6}$$

where $y_{ic}^s$ and $y_{ic}^t$ are the one-hot vectors for samples from SD and probability vectors predicted by the model for samples from TD, respectively.

### B. Self-training

Self-training is one of the earliest approaches in semi-supervised learning and has recently been widely used in DA [49]. The core idea is to assign pseudo labels to unlabeled samples for model training. By progressively incorporating pseudo-labeled TD samples into the training set, self-training can gradually correct domain shift.

The self-learning strategy is an iterative wrapper algorithm that starts by learning a supervised classifier on the labeled SD. Then, at each iteration, a proportion of the unlabeled TD data is selected based on confidence thresholds, and pseudo labels are assigned using the classifier's predictions.

Chen et al. [50] proposed a debiased self-training approach for semi-supervised training. They found out that training with pseudo labels aggressively in turn enlarges bias in certain categories. To alleviate this problem, they introduced a pseudo classification head $h_{\text{psd}}$, which is only optimized with pseudo labels generated from the original classification head $h_{\text{cls}}$. The parameters of the two heads are independent, ensuring that any errors in pseudo labels do not directly propagate bias into $h_{\text{cls}}$ during iterative self-training.

## III. Methodology

The flowchart of the proposed DHSNet is presented in Fig. 1, which consists of three main parts: feature extractor, domain adaptation, and self-training. The feature extractor processes samples from the SD and TD to extract transferable features. Domain adaptation aligns these features at the class level across domains. The self-training part utilizes predictions for TD samples to further optimize the model. Through the collaboration of these three parts, DHSNet can efficiently learn cross-scene invariant features while reducing the impact of erroneous pseudo-labels on the model.

As shown in Fig. 1, patches from labeled SD image and unlabeled TD image, denoted as $I \in \mathbb{R}^{ps \times ps \times c}$, where $ps$ is the patch size and $c$ is the number of spectral bands, are randomly selected (as shown by the black lines) and feed into the feature extractor. In the flowchart, purple lines represent labeled data flow, gray lines represent unlabeled data flow, and blue lines indicate the use of probability vectors. The extracted features from SD and TD are denoted as $\mathbf{z}^s$ and $\mathbf{z}^t$, respectively. These features are then input into the classification head $h_{\text{cls}}$ to get probability vectors $p^s$ and $p^t$, while the pseudo head $h_{\text{psd}}$ only get $\mathbf{z}^t$ and output predictions $p^{t'}$. The domain adaptation part applies the LMMD loss function for class-wise feature alignment between SD and TD. In the self-training process, pseudo labels from TD samples, generated by $h_{cls}$, are used to supervise the predictions from $h_{psd}$. Then parameters of both the feature extractor and $h_{psd}$ are optimized through gradient backpropagation. During testing, the well-trained feature extractor and classification head are used to classify samples from TD by extracting domain-invariant features.

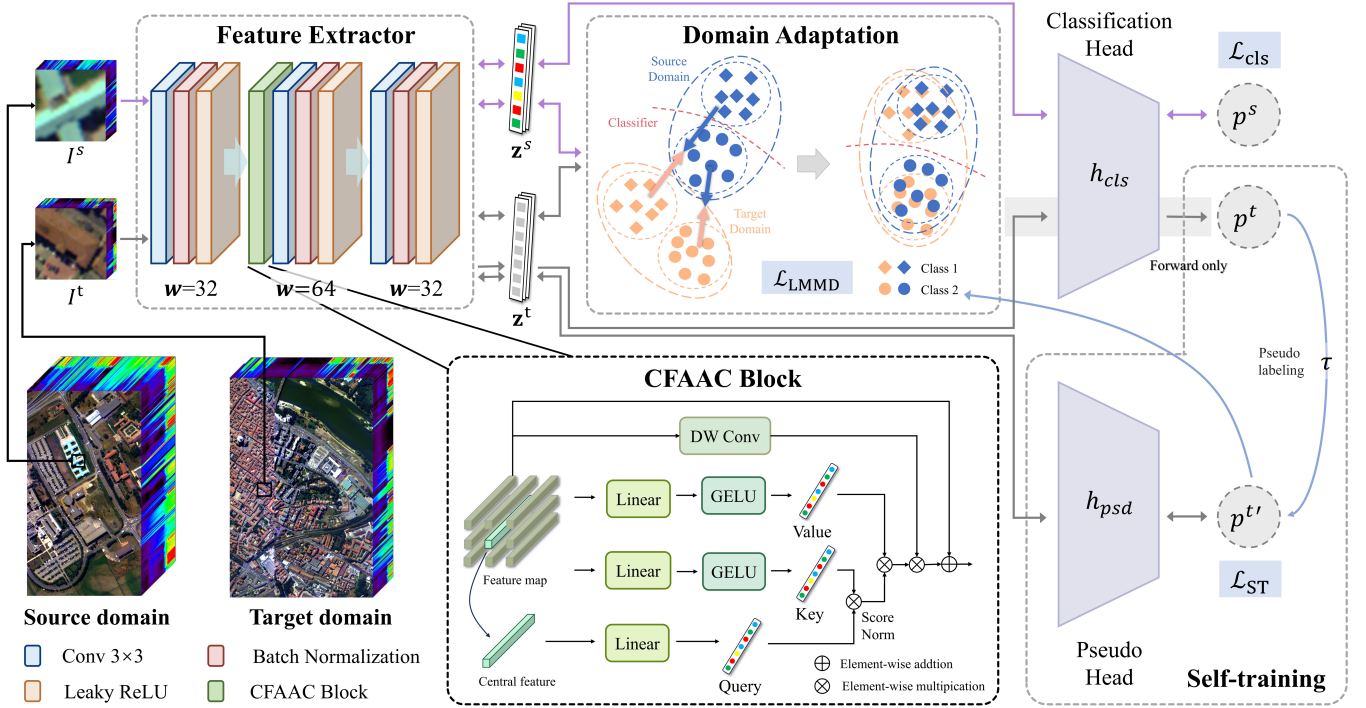

Fig. 1. Flowchart of the proposed DHSNet.

### A. Central Feature Attention Aware Convolution

To ensure our model accurately captures and represents hyperspectral cross-scene invariant features, a novel module called the central feature attention aware convolution block is introduced in the feature extractor. The module optimizes the use of features from the center of feature maps, which are more directly related to the classification of the original central pixels compared to peripheral ones. Additionally, the central features tend to aggregate more spectral information specific to certain land cover categories, making them less sensitive to the domain bias that can arise from incorporating more spatial information. It can serve as prior category information for the model through a gating mechanism. This module dynamically controls the convolution process, promoting a positive and efficient interaction between the extraction of spatial and spectral information.

The feature extractor is composed of three units, with the channel dimensions for each unit represented by $\boldsymbol{w}$ in the flowchart (Fig. 1). The first and third units follow a sequential structure, comprising a 3×3 convolution layer, batch normalization, and the leaky rectified linear unit (ReLU) activation function [51]. The second unit integrates a CFAAC block before these layers. The channel dimension in the second unit is set to 64, which is double that of the other units to ensure thorough feature extraction.

Details of the CFAAC block are also presented in the flowchart (Fig. 1). Inspired by the gated linear unit (GLU) [52], the CFAAC block consists of two streams: central feature attention and convolution. For unit $i$, the input feature map $F_i \in \mathbb{R}^{ps_i \times ps_i \times w_i}$ is first fed into two separate linear layers, followed by the Gaussian error linear unit (GELU) activation function, to generate the key matrix $K_i \in \mathbb{R}^{(ps_i \times ps_i) \times w_i}$ and the value matrix $V_i \in \mathbb{R}^{(ps_i \times ps_i) \times w_i}$. Simultaneously, the central pixel of $F_i$ is fed into another linear layer to generate a query vector $Q_i \in \mathbb{R}^{1 \times w_i}$. Each vector in $K_i$ computes its similarity with $Q_i$ via an inner product operation, determining the feature similarity in the latent space. These similarities are then normalized and applied as attention to each position. The attention matrix is multiplied by $V_i$ to obtain the result for the central feature attention stream.

In the convolution stream, spatial information already utilized in the central feature attention is processed using a depth-wise convolution, grouped by the channel dimension $w_i$ to enhance computational efficiency. Finally, the outputs from the two streams are element-wise multiplied and added to the original feature map $F_i$ using a residual design, producing the output of the CFAAC block. The specific formulations are as follows:

$$K_i = GELU\big(L_{key}(F_i)\big) \tag{7}$$

$$V_i = GELU\big(L_{value}(F_i)\big) \tag{8}$$

$$Q_i = L_{query}\big(Center(F_i)\big) \tag{9}$$

$$\text{CFAAC}(F_i) = \frac{Q_i^T K_i}{\sqrt{ps_i}} V_i \otimes \text{DWConv}(F_i) + F_i \qquad (10)$$

where $L(\cdot)$ represents a linear layer, and $Center(\cdot)$ refers to the extraction of the central pixel. The operator $\otimes$ denotes element-wise multiplication, and $ps_i$ is the patch size of $F_i$. DWConv$(\cdot)$ represents depth-wise convolution, a type of convolution operation where each input channel is convolved with a separate kernel.

The central feature attention mechanism differs from the scaled dot-product attention proposed by Vaswani et al. [53] by altering the many-to-many relationship between key vectors and query vectors. While both works measure pixel similarity, in our design, each pixel only queries the central one. This dynamic approach modulates the convolution process by giving more attention to features closer to the center. Another key distinction is the use of activation functions in our approach. It enhances the nonlinear representational capacity of the model and further differentiates our central feature attention from the scaled dot-product attention.

### B. Domain Adaptation

LMMD distance is applied in the domain adaptation part of DHSNet to align the feature distributions from the SD and TD that correspond to the same land cover class. In practice, the Gaussian kernel, which is a type of radial basis function, is utilized as the kernel function in LMMD. Since $\phi\ (\cdot)$ in formula (4) cannot be computed directly, we expand the square in the formula and leverage the properties of the kernel function in the RKHS to derive the following expression:

$$
\begin{aligned}
\mathcal{L}_{\text{LMMD}} = \frac{1}{C} \sum_{c=1}^{C} &[ \sum_{i=1}^{n_s} \sum_{j=1}^{n_s} w_i^{sc} w_j^{sc} r(\mathbf{z}_i^s, \mathbf{z}_j^s) \\
&+ \sum_{i=1}^{n_t} \sum_{j=1}^{n_t} w_i^{tc} w_j^{tc} r(\mathbf{z}_i^t, \mathbf{z}_j^t) \\
&- 2 \sum_{i=1}^{n_s} \sum_{j=1}^{n_t} w_i^{sc} w_j^{tc} r(\mathbf{z}_i^s, \mathbf{z}_j^t)]
\end{aligned}
\qquad (11)
$$

where the kernel $r(\cdot, \cdot)$ represents the inner product of the features in RKHS. The computation of $w_i^{sc}$ and $w_i^{tc}$ follow the same method as depicted in Formula (5) and Formula (6), respectively. Formula (11) is directly applied as the domain adaptation loss in DHSNet.

### C. Self-training

Inspired by debiased self-training, a dual classification head self-training strategy is introduced in DHSNet. Specifically, two classification heads with identical architectures are used independently: the classification head $h_{\text{cls}}$, which generates pseudo labels, and the pseudo classification head $h_{\text{psd}}$, which utilize these labels. As presented in the flowchart, $h_{\text{cls}}$ takes inputs $\mathbf{z}^s$ and $\mathbf{z}^t$, representing features from SD and TD, respectively, and outputs the land cover class probability vectors $p^s$ and $p^t$. Here, $p^s$, the classification result of the SD samples, is compared with the true labels, and cross-entropy loss function is employed to optimize the parameters of both the feature extractor and the classification head $h_{\text{cls}}$.

The cross-entropy loss is defined as:

$$\mathcal{L}_{\text{C-E}}(p_i, y_i) = - \sum_{c=1}^{C} y_i^c \log p_i^c \qquad (12)$$

where $y_i$ is the one-hot encoded labels, $p_i$ denotes the probabilistic predictions from the classification head $h_{\text{cls}}$, and $C$ is the number of classes. The classification loss for the SD is then defined as:

$$\mathcal{L}_{\text{cls}} = \frac{1}{n_s} \sum_{i=1}^{n_s} \mathcal{L}_{\text{C-E}}(h_{\text{cls}}(\mathbf{z}_i^s), y_i^s) \qquad (13)$$

where $n_s$ is the number of samples from SD.

The probabilistic predictions $p^t$ are then used as pseudo labels for the TD samples, but only for those with predicted probability for a specific class exceeds a threshold $\tau$ (empirically set to 0.95 in this paper). $h_{\text{psd}}$ processes only $\mathbf{z}^t$, and outputs their probabilistic predictions $p^{t\prime}$, which are used for self-training. The self-training loss for TD samples is formulated as:

$$\mathcal{L}_{\text{ST}} = \frac{1}{\hat{n}_t} \sum_{\|h_{\text{cls}}(\mathbf{z}_i^t)\|_{\infty} > \tau} \mathcal{L}_{\text{C-E}}\left( h_{\text{psd}}(\mathbf{z}_i^t), h_{\text{cls}}(\mathbf{z}_i^t) \right) \qquad (14)$$

where $\hat{n}_t$ is the number of TD samples that meet the aforementioned criteria of pseudo labels.

Integrating the above loss functions, the total training loss function of DHSNet is

$$\mathcal{L}_{total} = \mathcal{L}_{\text{cls}} + \lambda_{\text{LMMD}} \mathcal{L}_{\text{LMMD}} + \lambda_{\text{ST}} \mathcal{L}_{\text{ST}} \qquad (15)$$

where $\lambda_{\text{LMMD}}$ and $\lambda_{\text{ST}}$ are hyperparameters controlling the relative contributions of DA and self-training. Overall, the loss function of DHSNet consists of three parts. The classification loss for SD ensures that the model learns meaningful features for downstream tasks. The LMMD loss promotes class-wise feature alignment between SD and TD. The self-training loss extends the training data by incorporating pseudo-labeled TD samples, mitigating the model's bias towards SD.

## IV. EXPERIMENT

### A. Dataset

To evaluate the effectiveness of the proposed DHSNet, three publicly available cross-scene HSI datasets are used, including Houston, HyRANK, and Pavia.

TABLE I
LAND COVER CLASSES AND THE NUMBERS OF SAMPLES IN
THE HOUSTON DATASET

| No. | Class | Houston2013 | Houston2018 |
|---|---|---|---|
| C1 | Grass healthy | 345 | 1353 |
| C2 | Grass stressed | 365 | 488 |
| C3 | Trees | 365 | 2766 |
| C4 | Water | 285 | 22 |
| C5 | Residential buildings | 319 | 5347 |
| C6 | Non-residential buildings | 408 | 32459 |
| C7 | Road | 443 | 6365 |
| | Total | 2530 | 53200 |

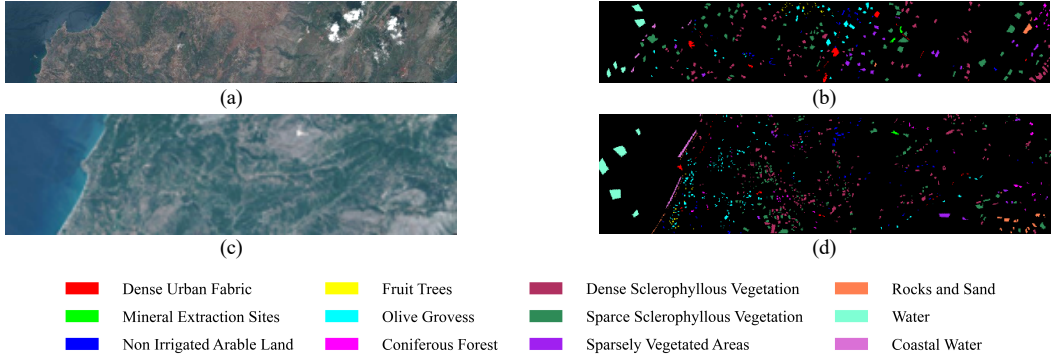

Fig. 2. Pseudo-color image and ground-truth map of HyRANK. (a) Pseudo-color image of Dioni. (b) Ground-truth map of Dioni. (c) Pseudo-color image of Loukia. (d) Ground-truth map of Loukia.

### TABLE II
### LAND COVER CLASSES AND THE NUMBERS OF SAMPLES IN THE HyRANK DATASET

| No. | Class | Dioni | Loukia |
|-----|-------|-------|--------|
| C1 | Dense Urban Fabric | 1262 | 206 |
| C2 | Mineral Extraction Sites | 204 | 54 |
| C3 | Non lrrigated Arable Land | 614 | 426 |
| C4 | Fruit Trees | 150 | 79 |
| C5 | Olive Groves | 1768 | 1107 |
| C6 | Coniferous Forest | 361 | 422 |
| C7 | Dense Sclerophyllous Vegetation | 5035 | 2996 |
| C8 | Sparce Sclerophyllous Vegetation | 6374 | 2361 |
| C9 | Sparsely Vegetated Areas | 1754 | 399 |
| C10 | Rocks and Sand | 492 | 453 |
| C11 | Water | 1612 | 1393 |
| C12 | Coastal Water | 398 | 421 |
| | Total | 20024 | 10317 |

sensors and at different times in Houston, TX, USA. Houston 2013 includes 349×1905 pixels and 144 spectral bands, whereas Houston 2018 includes 209×955 pixels and 48 spectral bands. To align with the Houston 2018 dataset, an overlapped region from Houston 2013 with 48 spectral bands and 209×955 pixels were selected. Both scenes share seven common land cover classes, with details shown in Table I and Fig. 3.

### TABLE III
### LAND COVER CLASSES AND THE NUMBER OF SAMPLES IN THE PAVIA DATASET

| No. | Class | Pavia University | Pavia Center |
|-----|-------|------------------|--------------|
| C1 | Trees | 3064 | 7598 |
| C2 | Asphalt | 6631 | 9248 |
| C3 | Bricks | 3682 | 2685 |
| C4 | Bitumen | 1330 | 7287 |
| C5 | Shadow | 947 | 2863 |
| C6 | Meadows | 18649 | 3090 |
| C7 | Bare soil | 5029 | 6584 |
| | Total | 39332 | 39355 |

2) *HyRANK:* This dataset is composed of two regions, Dioni (used as SD) and Loukia (used as TD). They are obtained from the Hyperion sensor (EO-1, USGS), featuring 176 spectral bands [56]. The sizes of Dioni and Loukia are 250×1376 and 249×945, respectively. There are 12 common classes between the two scenes, as detailed in Table II. The pseudo-color images and ground-truth maps for both scenes are shown in Fig. 2.

3) *Pavia:* This dataset includes Pavia University (PU, used as SD) [57] and Pavia Center (PC, used as TD) [58]. Both datasets were obtained using the Reflective Optics System Imaging Spectrometer (ROSIS) sensors in Pavia, northern Italy. PU contains 610×610 pixels and 103 spectral bands, whereas PC consists of 1096×1096 pixels and 102 spectral bands. For experimental purposes, PU and PC were cropped to 610 × 315 pixels and 1096 × 715 pixels, respectively. To ensure consistency, the last spectral band of PU was removed to match the number of spectral bands with PC. Both datasets feature seven common land cover classes, with detailed sample information in Table III. Fig. 4 displays their pseudo-color images and ground-truth maps.

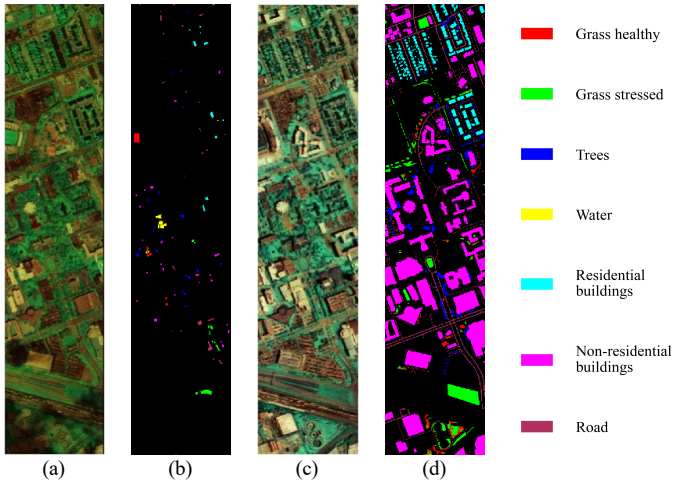

Fig. 3. Pseudo-color image and ground-truth map of Houston. (a) Pseudo-color image of Houston 2013. (b) Ground-truth map of Houston 2013. (c) Pseudo-color image of Houston 2018. (d) Ground-truth map of Houston 2018.

1) *Houston:* This dataset was obtained from the University of Houston campus and the neighboring urban area. It consists of two scenes, Houston 2013 [54] (used as SD) and Houston 2018 [55] (used as TD), which were captured with different

TABLE IV
QUANTITATIVE CLASSIFICATION RESULTS ON THE HOUSTON DATASET (TD: HOUSTON 2018)

| No. | Class | DAAN | DSAN | TSTnet | CLDA | DDAN-JCA | SDEnet | S2ECNet | DHSNet |
|-----|-------|------|------|--------|------|----------|--------|---------|--------|
| C1 | Grass healthy | 58.05±23.59 | 50.38±34.86 | 60.64±27.58 | <u>71.25±13.52</u> | 49.24±33.84 | 35.61±8.92 | 31.91±17.83 | **84.12±9.29** |
| C2 | Grass stressed | <u>81.84±1.89</u> | 73.04±16.88 | 69.50±5.70 | **86.26±4.76** | 15.07±19.00 | 63.81±22.60 | 71.61±4.65 | 79.23±2.93 |
| C3 | Trees | 64.39±4.97 | 63.22±12.16 | 65.63±3.16 | <u>67.52±4.79</u> | 56.96±15.55 | 64.99±6.26 | 58.26±13.65 | **71.76±2.82** |
| C4 | Water | 91.82±8.33 | 91.82±4.45 | **100.00±0.00** | 85.45±11.28 | 43.64±18.32 | 92.73±10.60 | **100.00±0.00** | <u>96.36±7.27</u> |
| C5 | Residential buildings | 77.26±9.58 | 63.76±12.69 | <u>87.33±3.37</u> | **93.16±4.17** | 76.45±18.58 | 72.52±9.24 | 61.34±3.48 | 85.32±5.65 |
| C6 | Non-residential buildings | 64.08±2.91 | 67.41±6.78 | **83.84±3.86** | 54.26± 7.02 | 73.61±10.25 | 60.28±5.85 | 50.81±6.69 | <u>83.80±4.32</u> |
| C7 | Road | 57.83±15.77 | 52.77±9.17 | 62.91±7.99 | **73.47±6.79** | 27.28±34.25 | 48.95±21.38 | <u>64.69±10.04</u> | 61.33±7.84 |
| | OA (%) | 66.16±3.90 | 65.16±4.14 | <u>78.84±2.61</u> | 64.55±4.22 | 61.48±1.40 | 60.11±5.24 | 55.37±3.36 | **80.23±1.92** |
| | AA (%) | 70.75±4.40 | 66.05±5.88 | <u>75.69±5.19</u> | 75.91± 2.54 | 48.89±11.41 | 62.70±8.07 | 62.66±3.62 | **80.27±1.41** |
| | Kappa × 100 | 50.29±5.38 | 47.47±4.49 | <u>65.87±3.81</u> | 51.65±4.27 | 35.88±9.21 | 41.46±9.01 | 38.87±2.68 | **68.52±2.07** |

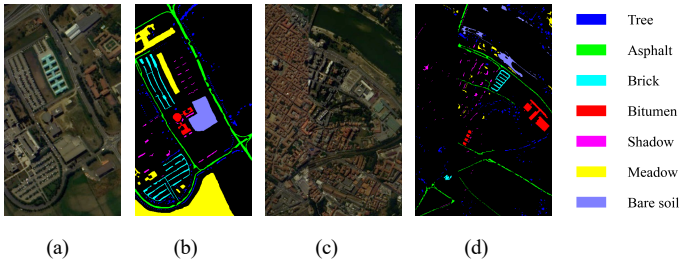

Fig. 4. Pseudo-color image and ground-truth map of Pavia. (a) Pseudo-color image of Pavia University. (b) Ground-truth map of Pavia University. (c) Pseudo-color image of Pavia Center. (d) Ground-truth map of Pavia Center.

### B. Implementation Details

Several state-of-the-art transfer learning algorithms were employed for comparison. For DA methods, we selected the following models: Dynamic Adversarial Adaptation Network (DAAN) [59], Deep Subdomain Adaption Network (DSAN) [48], TSTnet [43], CLDA [44] and DDAN-JCA [46]. In addition, for domain generalization (DG) methods, SDEnet [17] and S2ECNet [18] were selected for comparison.

1) For DAAN, the training consists of 200 epochs with a batch size of 40. The learning rate is $lr_w = (lr_0/(1 + \alpha w)^\beta)$, where $\alpha = 10$, $\beta = 0.75$, and $lr_0 = 0.01$. The optimizer is stochastic gradient descent (SGD) with a momentum of 0.9 and a weight decay of 0.0005. For all datasets, the input patch size is 5×5.

2) For DSAN, the training consists of 300 epochs with a batch size of 100, and the learning rate follows the same decay strategy as DAAN. The optimizer is SGD with a momentum of 0.9 and a weight decay of 0.0001. For all datasets, the input patch size is 5×5, and the parameter $\lambda$ is set to 0.5.

3) For TSTnet, the training consists of 500 epochs with a batch size of 100, and the learning rate decay strategy and optimizer settings are the same as DAAN. For all datasets, the input patch size is 13×13, with $\lambda_1$ and $\lambda_2$ set to 1 and 10 for Houston, and 1 and 0.1 for HyRANK and Pavia.

4) For CLDA, the training consists of 100 epochs with a batch size of 36. $lr_0$ is 0.0001, with $\alpha = 0.1$, $\beta = 0.01$, and $\gamma = 0.01$. All other settings are consistent with the method's open-source code. The input patch size is 5×5 for all datasets.

5) For DDAN-JCA, the training consists of 100 epochs with a batch size of 64. The parameter $\alpha$ is set to 0.5 for Pavia and 1 for Houston and HyRANK. The input patch size is 9×9 for Pavia and Houston, and 5×5 for HyRANK.

6) For SDEnet, the training consists of 400 epochs with a batch size of 100. The input patch size is 13 × 13 for all datasets, with $\lambda_1$, $\lambda_2$ and the embedding feature dimension $d_{se}$ set to 1, 0.1 and 64, respectively.

7) For S2ECNet, the training consists of 500 epochs with a batch size of 64. $lr_0$ is set to 0.001, and the Adam optimizer is used. zdim size is 256. Parameter $\lambda$ is 1. And the patch size is 13 × 13 for all datasets.

The experimental setup for DHSNet utilizes the Pytorch framework, employing the minibatch SGD for optimization. The momentum is set to 0.9 and $\ell_2$-norm regularization is set to 0.0001. The learning rate and batch size for all three datasets are set to 0.01 and 100, respectively. For the feature extractor, the convolution layers and batch normalization layer use the Kaiming distribution, and initialized with a constant value of 1. The learning rate adjustment follows the formula: $lr_w = (lr_0/(1 + \alpha w)^\beta)$, where $w$ indicates the training progress linearly ranging from 0 to 1, $lr_0$ is the initial learning rate, $\alpha = 10$, and $\beta = 0.75$. The number of training epochs is 200. Patch sizes for Houston, HyRANK and Pavia are empirically set to 15×15, 7×7 and 9×9.

The classification performance is quantitatively evaluated using the overall accuracy (OA), the average accuracy (AA) and the Kappa coefficient [60]. To ensure unbiased results, all experiments are repeated five times independently, with the mean values of the evaluation metrics reported for comprehensive analysis.

### C. Experimental Results

Tables IV-VI display the class-specific classification accuracy, OA, AA, and Kappa coefficients for the aforementioned methods across three datasets. Bold indicates the highest accuracy, and underline indicates the second-best accuracy. And visual classification results are further shown in Figs.5-7. In these maps, labeled pixels are displayed as ground truth and unlabeled pixels as backgrounds. From the quantitative results, we can see:

1) Outstanding quantitative performance: Compared to other methods, the proposed DHSNet achieves the best performance

TABLE V
QUANTITATIVE CLASSIFICATION RESULTS ON THE HYRANK DATASET (TD: LOUKIA)

| No. | Class | DAAN | DSAN | TSTnet | CLDA | DDAN-JCA | SDEnet | S2ECNet | DHSNet |
|---|---|---|---|---|---|---|---|---|---|
| C1 | Dense Urban Fabric | 30.29±7.43 | 28.54±17.53 | _43.69±22.73_ | 28.93±27.35 | 6.70±1.12 | 33.40±12.23 | 27.28±11.87 | **64.37±3.08** |
| C2 | Mineral Extraction Sites | **87.04±9.59** | _81.11±19.68_ | 38.89±47.66 | 79.63±39.82 | 30.37±37.27 | 47.41±27.31 | 62.22±30.13 | 75.56±11.26 |
| C3 | Non Irrigated Arable Land | 37.32±13.09 | 65.12±22.93 | 47.93±16.98 | **82.35±6.34** | 56.95±15.74 | 40.38±18.99 | 35.12±10.04 | _75.49±9.47_ |
| C4 | Fruit Trees | 73.16±27.56 | 60.51±28.52 | 75.70±12.05 | **99.75±0.51** | 67.85±12.92 | 20.51±20.65 | 49.37±23.18 | _81.01±4.67_ |
| C5 | Olive Groves | 17.16±16.30 | _23.43±27.51_ | 9.65±12.50 | 3.41±2.97 | 5.35±3.22 | 3.58±5.66 | 16.68±13.03 | **47.90±16.70** |
| C6 | Coniferous Forest | 51.71±17.34 | 36.73±27.87 | 25.92±13.08 | 53.79±5.91 | 47.87±20.35 | **55.55±9.14** | 44.12±6.71 | _54.27±9.06_ |
| C7 | Dense Sclerophyllous Vegetation | 67.46±6.86 | 61.79±3.94 | 71.68±7.50 | 66.56±4.85 | 63.55±9.26 | 66.7±13.52 | **82.42±4.29** | _75.55±5.42_ |
| C8 | Sparce Sclerophyllous Vegetation | 49.14±14.24 | 56.73±14.39 | _59.88±11.29_ | 21.75±11.07 | 24.37±5.63 | 52.6±11.29 | 41.33±10.11 | **67.62±6.90** |
| C9 | Sparsely Vegetated Areas | 29.42±20.12 | 46.42±28.68 | 50.03±21.62 | 45.31±24.72 | 22.81±9.90 | 41.15±15.11 | _62.71±12.57_ | **77.99±6.14** |
| C10 | Rocks and Sand | 24.15±21.43 | 33.16±11.65 | 23.58±24.70 | _53.95±27.16_ | 50.29±23.75 | 15.14±13.98 | 16.34±24.55 | **57.00±21.77** |
| C11 | Water | 97.47±2.61 | 94.93±8.11 | _99.77±0.46_ | 100.00±0.00 | 95.03±6.80 | 100.00±0.00 | 100.00±0.00 | 100.00±0.00. |
| C12 | Coastal Water | 98.19±2.21 | 77.58±37.77 | 100.00±0.00 | 100.00±0.00 | _99.95±0.10_ | 100.00±0.00 | 100.00±0.00 | 99.71±0.38 |
| | OA (%) | 57.32±3.74 | 58.32±5.13 | _60.77±2.03_ | 53.73±2.47 | 49.73±3.81 | 56.64±3.10 | 60.41±2.60 | **73.28±2.28** |
| | AA (%) | 55.21±2.50 | 55.50±6.25 | 53.89±4.76 | _61.29±5.40_ | 47.59±3.90 | 48.03±5.57 | 53.13±2.60 | **73.04±1.14** |
| | Kappa × 100 | 49.75±3.98 | 50.75±6.00 | _52.54±1.84_ | 47.23±-2.70 | 41.86±3.56 | 47.76±3.38 | 51.90±3.51 | **68.04±2.46** |

TABLE VI
QUANTITATIVE CLASSIFICATION RESULTS ON THE PAVIA DATASET (TD: PAVIA CENTER)

| No. | Class | DAAN | DSAN | TSTnet | CLDA | DDAN-JCA | SDEnet | S2ECNet | DHSNet |
|---|---|---|---|---|---|---|---|---|---|
| C1 | Trees | 89.60±11.43 | 87.72±8.48 | 87.77±8.06 | 94.95±3.41 | **97.66±1.48** | 87.35±4.17 | 86.72±5.64 | _94.81±2.96_ |
| C2 | Asphalt | 82.64±11.36 | 87.94±2.60 | _92.55±3.82_ | **99.86±0.10** | 92.54±2.24 | 74.18±4.16 | 76.29±3.78 | 97.13±0.82 |
| C3 | Bricks | **87.99±7.29** | 80.53±21.40 | 75.86±6.94 | 76.60±3.85 | 77.91±6.60 | 66.00±9.45 | 76.78±12.93 | _84.27±5.43_ |
| C4 | Bitumen | 61.61±28.87 | 65.57±3.71 | 0.97±0.34 | 71.80±24.43 | 22.14±21.66 | 84.66±1.38 | _85.21±0.67_ | **86.41±1.83** |
| C5 | Shadow | 80.27±6.29 | 89.05±6.12 | 97.23±1.37 | _99.99±0.01_ | 99.66±0.53 | 85.18±7.23 | 80.13±6.65 | **100.00±0.00** |
| C6 | Meadows | 47.92±15.28 | 65.18±19.13 | 77.18±9.70 | **85.88±9.87** | 73.37±12.97 | 76.19±4.26 | 73.68±8.06 | _84.76±6.97_ |
| C7 | Bare soil | 78.32±8.60 | 77.81±5.21 | 86.65±5.41 | **92.72±2.66** | 78.30±8.00 | 78.23±10.26 | 82.78±6.48 | _91.09±1.37_ |
| | OA (%) | 76.83±4.92 | 79.85±2.48 | 71.68±1.45 | _89.85±4.62_ | 76.13±3.95 | 79.74±1.83 | 81.15±1.66 | **92.05±0.31** |
| | AA (%) | 75.48±3.44 | 79.11±3.41 | 74.03±1.98 | _88.83±3.85_ | 77.37±2.15 | 78.83±1.63 | 80.23±1.74 | **91.21±0.97** |
| | Kappa × 100 | 72.13±5.87 | 75.76±3.02 | 66.64±1.69 | _87.72±5.65_ | 71.16±4.79 | 75.73±2.17 | 77.43±1.99 | **90.42±0.38** |

in terms of class-specific accuracy, OA, AA, and Kappa coefficient. For the Houston dataset, DHSNet's OA exceeds 80.00%, outperforming TSTnet, which has a high performance on this dataset at 78.84%, and leading by a large margin over all other methods. For the HyRANK dataset, where vegetation classes are prone to misclassification and most methods have an OA just above 60.00%, DHSNet's OA breaks through 70%. For the Pavia dataset, DHSNet slightly leads over CLDA (89.95%), which performs well on this dataset and substantially outperforms the rest of the methods.

2) Excellent class balance: Through class-wise feature alignment, DHSNet can obtain superior and balanced class-specific accuracy for some difficult-to-distinguish categories. For instance, in the Houston dataset, DHSNet achieves nearly equal classification accuracy (about 80%) for both the first class (Grass healthy) and the second class (Grass stressed), indicating minimal class imbalance. In the HyRANK dataset, DHSNet significantly outperforms other methods in classifying Olive Groves, a class for which other methods achieve less than 25% accuracy. DHSNet achieves an average accuracy of 47.90% for this class, though with some room for improvement in stability.

3) Deep transfer learning baseline models like DAAN and DSAN, not specifically designed for HSI, perform reasonably well across the datasets, particularly in cases with many classes (such as the HyRANK dataset). However, these methods reach a performance ceiling compared to HSI-specific methods, which incorporate additional mechanisms such as pseudo labels or topological information.

4) DA and DG methods show strengths on different datasets.

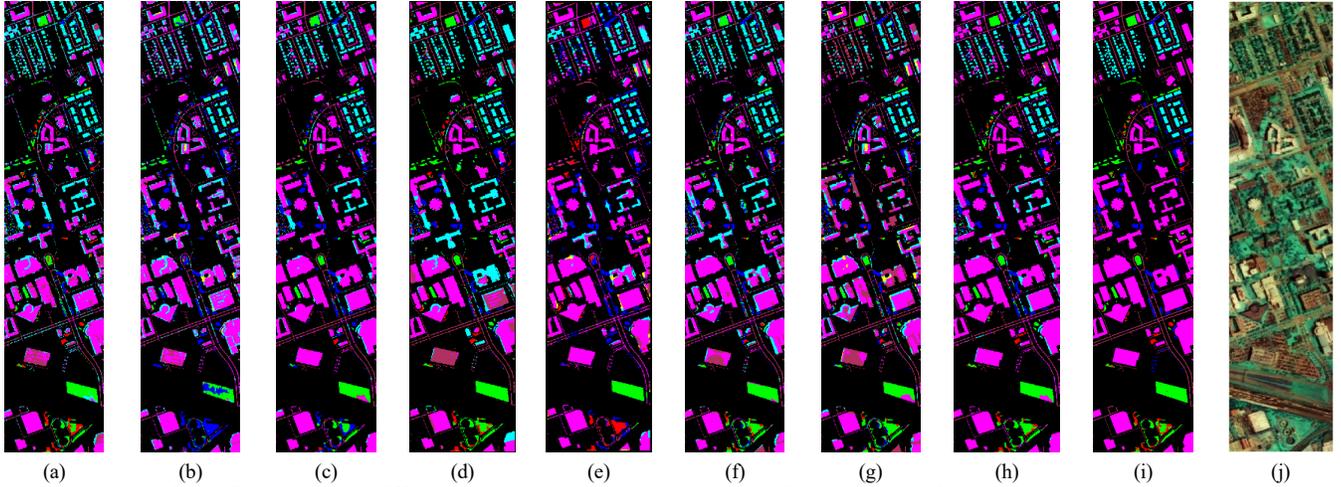

Fig. 5. Visualization of classification maps by different methods on the Houston dataset, including (a) DAAN (68.73%), (b) DSAN (64.11%), (c) TSTnet (79.64%), (d) CLDA (69.06%), (e) DDAN-JCA (67.02%), (f) SDEnet (69.13%), (g) S2ECNet (59.15%), (h) DHSNet (81.27%), (i) Ground truth, (j) Pseudo-color image of Houston 2018.

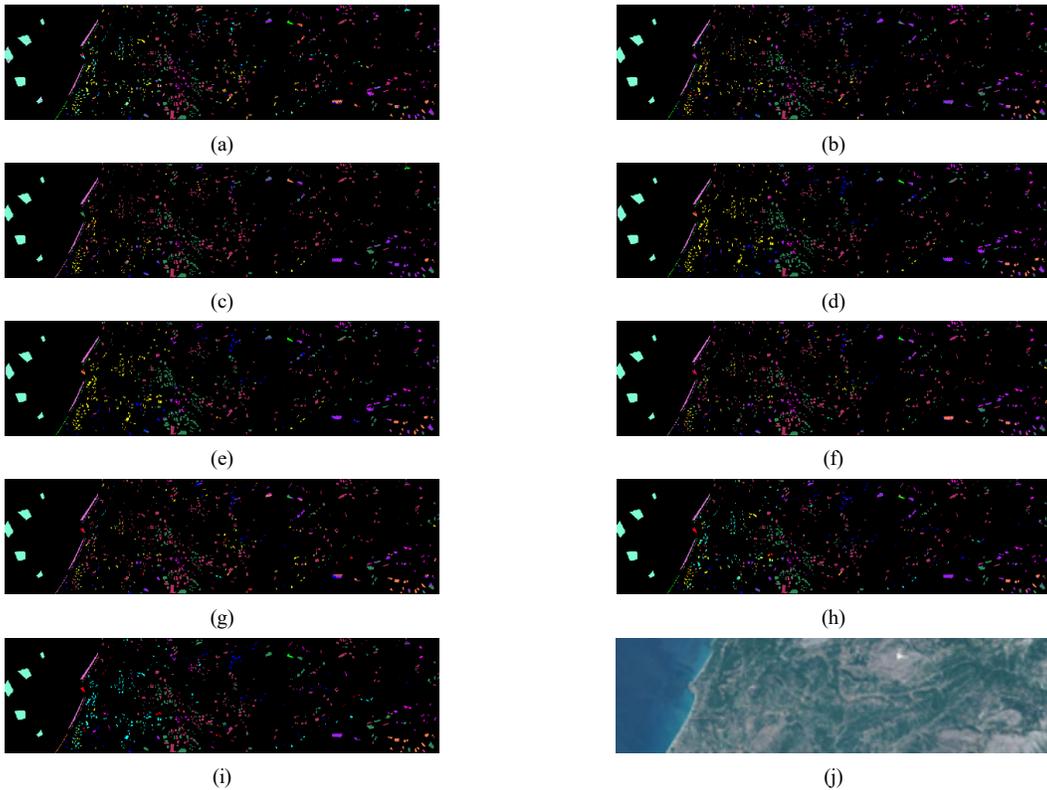

Fig. 6. Visualization of classification maps by different methods on the HyRANK dataset, including (a) DAAN (57.85%), (b) DSAN (64.01%), (c) TSTnet (61.65%), (d) CLDA (59.10%), (e) DDAN-JCA (59.01%), (f) SDEnet (57.47%), (g) S2ECNet (62.28%), (h) DHSNet (73.50%), (i) Ground truth, (j) Pseudo-color image of Loukia.

For the Houston dataset, DG methods (SDEnet and S2ECNet) perform poorly compared to DA methods, but on the Pavia dataset, DG methods surpass all DA methods except CLDA and DHSNet. DG methods appear to have better sample expansion capabilities in large-scale urban scenarios like Pavia, whereas DA methods excel across more challenging and diverse datasets like HyRANK. The performance of DG methods seems to be more dependent on the number of available samples, which could explain their underperformance in this study, where a fixed number of samples per class was used. In their original studies, DG methods used a proportional sampling strategy, leading to a greater number of samples compared to the fixed sample size used in some datasets here, allowing them to perform better.

### D. Ablation Study

The ablation study is designed to demonstrate the validity of CFAAC block, domain adaptation, and self-training and evaluate their contributions. We adopt five methods on three datasets. To verify the effectiveness of the CFAAC block, a baseline model using the same feature extractor architecture as DHSNet but without the CFAAC block is established. Besides, only the feature extractor of DHSNet is retained for this comparison. Besides, we separately add the domain adaptation and self-training components to the feature extractor to validate their respective effectiveness. Experiment designators for each condition are illustrated in Table VII.

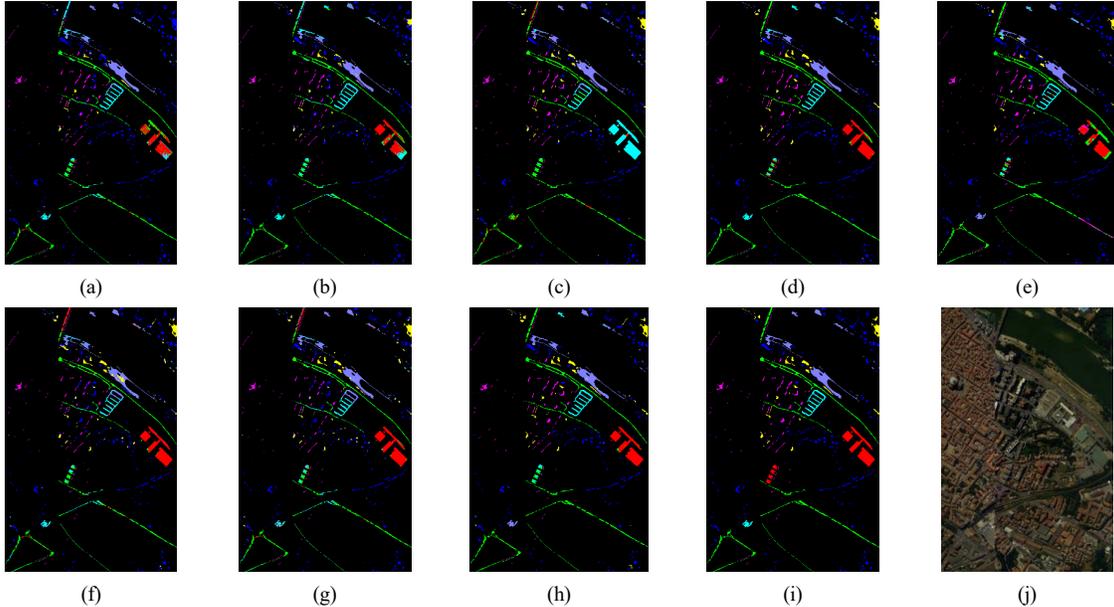

Fig. 7. Visualization of classification maps by different methods on the Pavia dataset, including (a) DAAN (78.51%), (b) DSAN (81.28%), (c) TSTnet (71.79%), (d) CLDA (93.13%), (e) DDAN-JCA (83.67%), (f) SDEnet (80.56%), (g) S2ECNet (82.36%), (h) DHSNet (93.55%), (i) Ground truth, (j) Pseudo-color image of Pavia Center.

TABLE VII
ABLATION COMPARISON OF MODULES OF DHSNET

| Method | | baseline | FE only | FE+LMMD | FE+ST | FE+LMMD+ST |
|---|---|---|---|---|---|---|
| Houston | OA(%) | 77.86±2.10 | 78.54±1.24 | 76.45±0.93 | <u>79.17±1.82</u> | **79.25±2.07** |
| | AA(%) | 74.87±1.39 | <u>76.92±2.25</u> | 73.76±2.79 | 73.86±1.20 | **78.14±4.12** |
| | Kappa×100 | 62.85±3.45 | <u>65.35±1.42</u> | 61.89±2.21 | 64.64±2.31 | **67.02±3.65** |
| HyRANK | OA(%) | 63.52±2.17 | 64.13±4.06 | <u>70.56±1.85</u> | 65.30±2.02 | **71.25±1.11** |
| | AA(%) | 58.99±2.83 | 56.30±1.86 | <u>72.58±2.48</u> | 63.42±4.05 | **72.90±1.91** |
| | Kappa×100 | 55.96±2.29 | 56.82±4.48 | <u>64.80±1.89</u> | 58.31±2.38 | **65.74±1.31** |
| Pavia | OA(%) | 81.02±2.47 | 81.72±3.33 | <u>91.34±0.54</u> | 89.84±1.67 | **92.05±0.31** |
| | AA(%) | 80.64±1.83 | 79.81±6.46 | <u>90.29±0.28</u> | 87.57±1.45 | **91.21±0.97** |
| | Kappa×100 | 77.02±3.00 | 77.89±4.07 | 89.57±0.65 | <u>87.73±2.00</u> | **90.42±0.38** |

As shown in Table VII, FE with CFAAC block achieves approximately a 0.70% improvement in OA across the three datasets compared to the baseline. This demonstrates the effectiveness of the CFAAC block in extracting and representing domain-invariant features for HSI. Except for the Houston dataset, incorporating only the LMMD loss significantly enhances the OA, with the HyRANK dataset showing an improvement of approximately 6.00% and the Pavia dataset about 10.00%. Applying the self-training strategy alone results in an OA improvement of around 1% for the Houston and HyRANK datasets, while the Pavia dataset sees an increase of over 8.00%. Furthermore, the combined application of LMMD and self-training achieves the best performance across all datasets, outperforming the individual application of each component. This indicates that the combination of LMMD and self-training is effective for cross-scene HSI classification.

### E. Validity of Pseudo Classification Head

To demonstrate the validity of the pseudo classification head, a series of experiments using self-training, with and without the pseudo classification head, were implemented under various experimental conditions. The experimental settings are shown in Table VIII. It is noteworthy that all these experiments use the same FE as DHSNet and the same pseudo label criterion threshold $\tau$. The results are presented in Fig. 9.

In all datasets, regardless of whether LMMD is applied or not, employing a pseudo classification head in the self-training strategy consistently outperforms the approach without it. This improvement is particularly pronounced on Pavia dataset and followed by HyRANK dataset. Therefore, the effectiveness of the pseudo classification head in self-training has been successfully demonstrated.

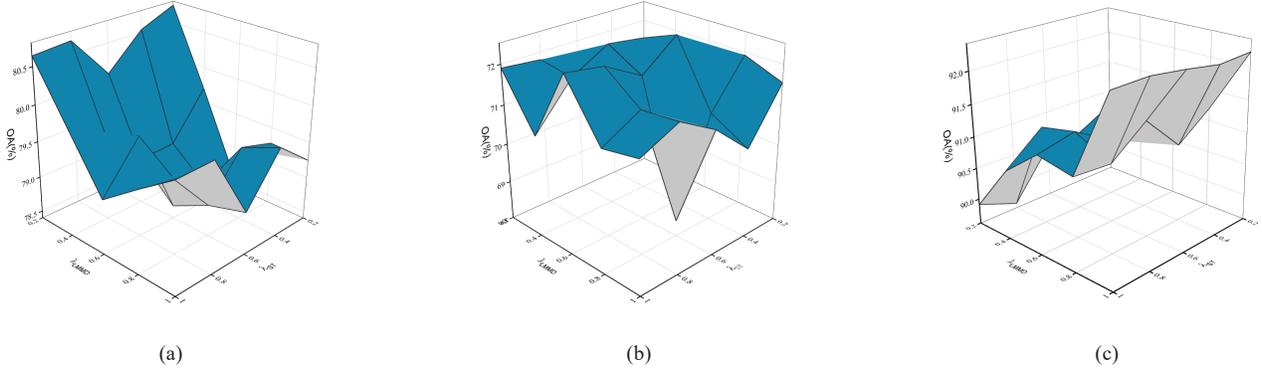

(a)                                    (b)                                    (c)

Fig. 8. OA (%) of different combinations of $\lambda_{\text{LMMD}}$ and $\lambda_{\text{ST}}$. (a) Houston, (b) HyRANK, (c) Pavia.

TABLE VIII
EXPERIMENT SETTINGS TO DEMONSTRATE THE VALIDITY OF
PSEUDO CLASSIFICATION HEAD

| Experiment | Pseudo Head | LMMD |
|------------|-------------|------|
| a | ✗ | ✗ |
| b | ✓ | ✗ |
| c | ✗ | ✓ |
| d | ✓ | ✓ |

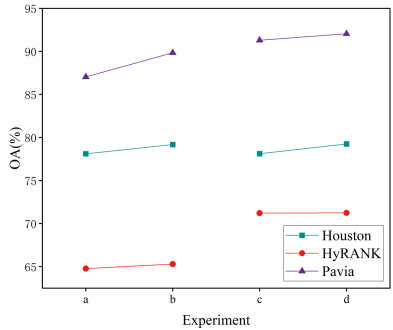

Fig. 9. OA (%) of different experimental settings with pseudo head and
LMMD on the three datasets.

### F. Parameter Sensitivity Analysis

There are three parameters in DHSNet, i.e., patch size, $\lambda_{\text{LMMD}}$, and $\lambda_{\text{ST}}$. Patch size controls the size of the input HSI data cube, $\lambda_{\text{LMMD}}$ and $\lambda_{\text{ST}}$ determine the weight of LMMD and self-training in the total loss function. We initially empirically set both $\lambda_{\text{LMMD}}$ and $\lambda_{\text{ST}}$ to 1 and searched for the optimal patch size for each dataset. Patches are selected from $\{5 \times 5, 7 \times 7, 9 \times 9, 11 \times 11, 13 \times 13, 15 \times 15, 17 \times 17\}$ for verification.

As shown in Fig. 10, the best patch size of Houston, HyRANK and Pavia are 15, 7 and 11 respectively. HyRANK dataset is most sensitive to patch size, which experienced a significant increment in OA of over 30% from $5 \times 5$ to $7 \times 7$, and then continued to decline as the patch size increased. Pavia dataset is the least sensitive to patch size, with the range of OA not exceeding 2% across different patch sizes.

In the following discussion, patch size is set to the optimal for the corresponding dataset. The value ranges of $\lambda_{\text{LMMD}}$ and $\lambda_{\text{ST}}$ are both set $\{0.2, 0.4, 0.6, 0.8, 1\}$. Fig. 8 presents the changing trend of OA in three datasets with different $\lambda_{\text{LMMD}}$ and $\lambda_{\text{ST}}$ combinations. As shown in Fig. 8, DHSNet is more sensitive to $\lambda_{\text{LMMD}}$ than $\lambda_{\text{ST}}$. This is consistent with the contributions to the model's performance revealed by the ablation study when considering each part separately. OA of Houston dataset generally decreases with the increase of $\lambda_{\text{LMMD}}$, while Pavia shows the opposite trend. HyRANK exhibits an increasing and then decreasing trend on $\lambda_{\text{LMMD}}$, and it is more volatile for $\lambda_{\text{ST}}$. Overall, the sensitivity to these two parameters, from least to most, is Pavia, Houston, and HyRANK.

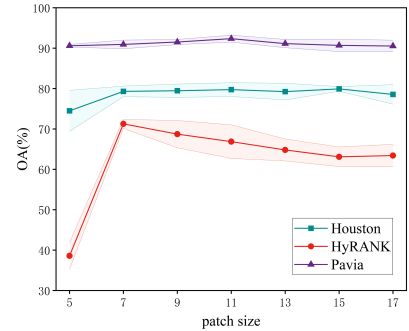

Fig. 10. OA (%) of different patch size on the three datasets. The shaded area represents the standard deviation.

Based on our experiments, the optimal $\lambda_{\text{LMMD}}$ and $\lambda_{\text{ST}}$ combinations for Houston, HyRANK and Pavia are $\{0.2, 0.2\}$, $\{0.6, 0.4\}$ and $\{1, 0.8\}$ respectively. Notably, the discrepancies in OA obtained from these diverse parameter configurations are predominantly constrained within a 2% interval, substantiating the stability and robustness of our proposed model.

### G. Design of CFAAC block

To design a module that can enhance the capacity of domain-invariant feature capturing and representation, we have explored four different implementations of the CFAAC block as depicted in Fig. 11. To amplify the weight calculated by the similarity of key and query, we abandoned the use of the softmax function. Correspondingly, to compensate for the loss of non-linear representation capability in the CFAAC block due to this approach, we incorporated the GELU activation function into the block. The main difference between the four implementations is the position of the activation function.

With other parameters empirically set, the experimental results obtained on the Pavia dataset are shown in Table IX.

The optimal performance was achieved when the activation function was placed after both the key and value operations. This configuration attained the highest OA, AA, and Kappa coefficient among the four experimental groups, with OA reaching 90.27%. The performance of placing the activation function after the depth-wise convolution and after the key operation was similar, both lagging behind the optimal experimental group by approximately 0.2%. The configuration without any activation function performed the worst, with an OA of 89.64%. These results demonstrate the necessity of the activation function in the CFAAC block.

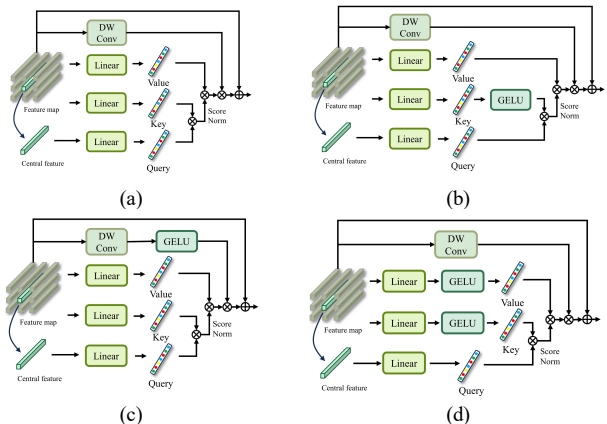

Figure. 11. Different implementations of CFAAC block, including: a) without activation function, b) after key, c) after deep-wise convolution, d) the default design of CFAAC block used in DHSNet, with activation functions after key and value.

TABLE IX
PERFORMANCE OF DIFFERENT BLOCK DESIGNS

| Design | a | b | c | d |
|---|---|---|---|---|
| OA(%) | 89.64±0.86 | 90.05±0.94 | 90.09±0.91 | **90.27±0.89** |
| AA(%) | 90.03±1.01 | 90.49±1.11 | 90.37±1.04 | **90.81±1.09** |
| Kappa ×100 | 87.54±1.04 | 88.04±1.12 | 88.08±1.09 | **88.30±1.07** |

## V. CONCLUSION

In this article, we propose an efficient network, DHSNet, for HSI cross-scene classification. The framework integrates class-wise domain adaptation and self-training to collaboratively achieve fine-grained feature alignment and mitigate the model's training bias on source domain. Additionally, a novel central feature attention aware convolution block is incorporated into our feature extractor, demonstrating effectiveness in extracting cross-scene invariant features and showing potential for other HSI cross-scene classification methods. Comprehensive experiments and analyses on three cross-scene HSI datasets suggest that DHSNet outperforms state-of-the-art methods. In future work, we will concentrate on addressing the category classification accuracy gap, exploring more detailed improvements in domain adaptation and self-training.